\documentclass[letterpaper, 10 pt, journal, twoside]{IEEEtran}

\usepackage{mystyle}

\newcommand{\proposedMappingFramework}{UFOMap}

\title{\LARGE \bf
{\proposedMappingFramework{}: An Efficient Probabilistic 3D Mapping \\ Framework That Embraces the Unknown}
}

\author{Daniel Duberg and Patric Jensfelt

\thanks{The authors are with the Division of Robotics, Perception and Learning (RPL), KTH Royal Institute of Technology, Stockholm, Sweden.
        {\tt\footnotesize \{dduberg, patric\}@kth.se}}%
}

\begin{document}

\maketitle
\thispagestyle{empty}
\pagestyle{empty}

\begin{abstract}
	3D models are an essential part of many robotic applications. In applications where the environment is unknown a-priori, or where only a part of the environment is known, it is important that the 3D model can handle the unknown space efficiently. Path planning, exploration, and reconstruction all fall into this category. In this paper we present an extension to OctoMap which we call \proposedMappingFramework{}. \proposedMappingFramework{} uses an explicit representation of all three states in the map, i.e., occupied, free, and unknown. This gives, surprisingly, a more memory efficient representation. Furthermore, we provide methods that allow for significantly faster insertions into the octree. This enables real-time colored volumetric mapping at high resolution (below 1~cm). \proposedMappingFramework{} is contributed as a C++ library that can be used standalone but is also integrated into ROS.
\end{abstract}

\section{Introduction}
\label{sec:introduction}
Many robot tasks require a 3D model of the environment to be completed. For navigation and manipulation tasks the model is often used to calculate collision free paths and in exploration to determine where new information can be found and how to get to it. 3D SLAM updates the model and localizes the robot using it.

A number of different methods for modeling 3D environments exists: point clouds, elevation maps~\cite{kweon1989terrain}, multi-level surface maps~\cite{triebel2006multi}, octrees~\cite{meagher1982geometric}, and signed distance fields~\cite{Oleynikova2016SignedPlanning}. All of these methods are able to represent occupied and free space. However, only octrees and signed distance fields are able to represent the unknown space. Moreover, point clouds, elevation maps, and multi-level surface maps are not able to represent arbitrary 3D environments.

\begin{figure}
    \centering
    \includegraphics[width=\linewidth]{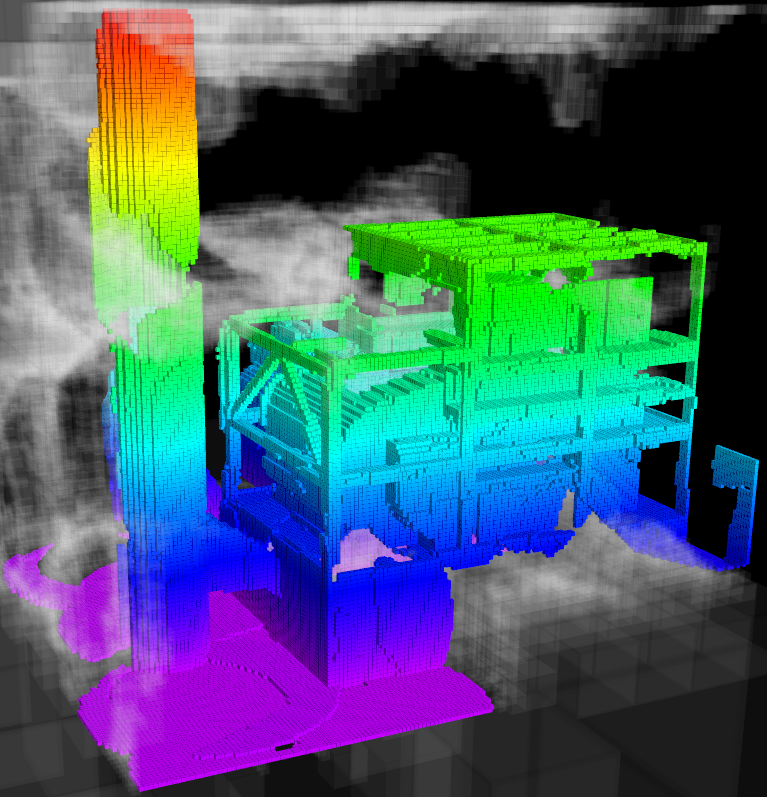}
    \caption{\proposedMappingFramework{} representation of an in-progress exploration of a power plant. The colored voxels are occupied space. The white voxels are unknown space which are represented explicitly in \proposedMappingFramework{}.}
    \label{fig:introduction}
\end{figure}

\begin{figure*}
    \centering
    \subfloat[]{
        \includegraphics[width=0.32\linewidth]{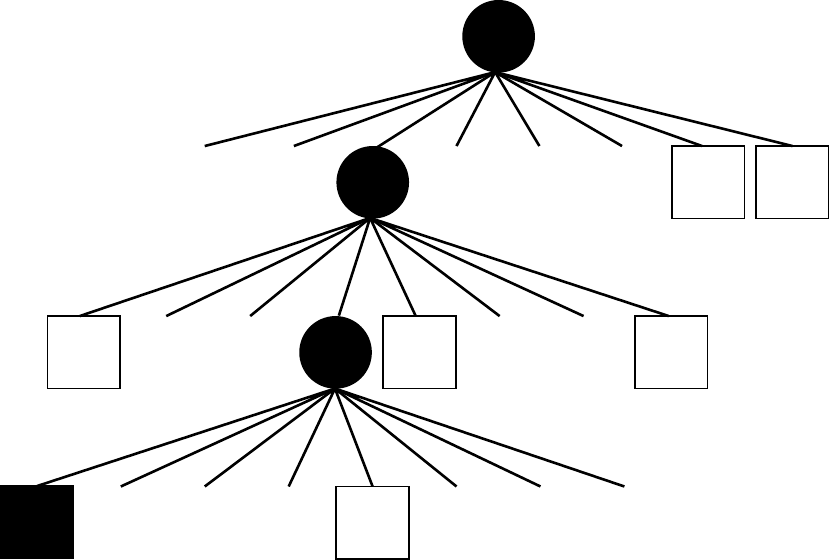}
        \label{fig:octomap_tree}
    }
    \subfloat[]{
        \includegraphics[width=0.32\linewidth]{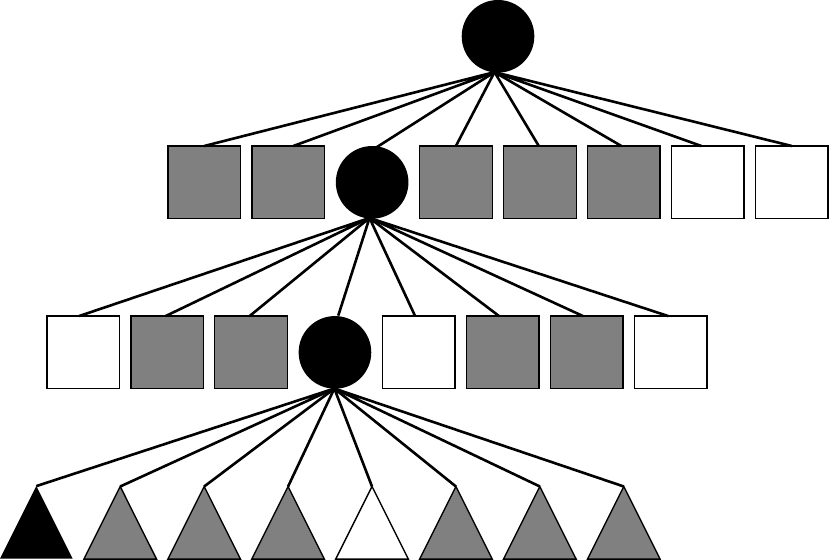}
        \label{fig:ufomap_tree}
    }
    \subfloat[]{
        \includegraphics[width=0.32\linewidth]{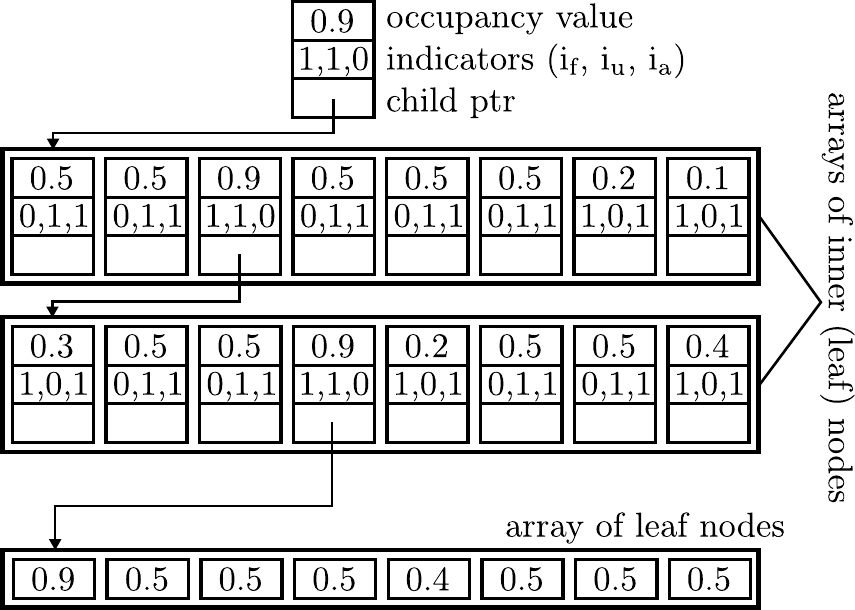}
        \label{fig:ufomap_datastructure}
    }
    \caption{Example of an octree. The OctoMap tree representation is shown in \protect\subref{fig:octomap_tree} and the corresponding tree representation in \proposedMappingFramework{} is displayed in \protect\subref{fig:ufomap_tree}. Black indicates occupied space, white free space, and grey unknown space. Circles means that it is an inner node, square an inner leaf node, and triangle a leaf node. Note that OctoMap does not make a distinction between leaf nodes types. \protect\subref{fig:ufomap_datastructure} shows how \proposedMappingFramework{} stores that octree in memory.}
    \label{fig:tree_structure}
\end{figure*}

OctoMap~\cite{hornung2013octomap}, one of the most popular mapping frameworks, is based on octrees. It provides a readily available, reliable, and efficient 3D mapping framework. OctoMap provides researchers, also those not focusing on mapping, an off the shelf solution for mapping. While OctoMap excels in many use cases, there exist certain areas where it can be improved. In this paper we focus on two areas.

First, in OctoMap, unknown space is not modeled explictly as occupied and free space. In algorithms where the unknown space is used extensively, OctoMap's implicit representation of unknown space can be a bottleneck. Use cases where this can be a problem are, for example, collision checking, path planning where we want to know if it is possible for a robot to move from one location to another and next-best view exploration methods such as \cite{bircher2016receding}, \cite{selin2019efficient}, \cite{Barbosa2019GuidingLogic}, and \cite{Schmid2020AnEnvironments}.
In the first two cases the common workaround is to treat unknown space as free space \cite{Pivtoraiko2013IncrementalEnvironments, Chen2016OnlineEnvironments}. This way, only occupied space has to be considered in the collision checking. However, this increases the probability of collisions. Regarding the third use case, \cite{Schmid2020AnEnvironments} states that "\emph{computing volumetric information gains can account for up to 95\% of a planner’s run-time.}" The information gain for a certain sensor/robot pose in an exploration scenario is typically calculated as a function of the unknown space. In \cite{selin2019efficient} the information gain is approximated, leading to a reduced need to access unknown space in the OctoMap and it is one of the major contributing factors of the performance improvement over the baseline \cite{bircher2016receding}.

Second, manipulating the content of an OctoMap is time consuming and limited. For example, inserting a single point cloud takes on the order of a second. This means that real-time volumetric mapping with an OctoMap is prohibitively slow in many applications. Furthermore, when inserting or deleting data in the map (e.g., mapping) you cannot also at the same time iterate over it to extract information (e.g. exploration). In many use cases the solution is to make a copy of the entire OctoMap structure and let consumers operate on it during the update.

In this paper we present an extension to OctoMap taking these shortcomings into account. The contributions of this work are as follows:
\begin{itemize}
    \item We present a memory efficient 3D mapping framework based on octrees, called \proposedMappingFramework{}, that explicitly represents the unknown space, together with occupied and free space.
    \item We introduce different methods of incorporating data into the octree that are significantly faster than OctoMap's. This enables real-time colored mapping at a higher resolution.
    \item We improve the overall efficiency compared to OctoMap and allow for inserting/deleting data and iterate over the octree at the same time.
    \item As opposed to OctoMap, most functions in \proposedMappingFramework{} can work at different resolutions of the octree. Making it possible for different algorithms with varying requirements to use the same map representation.
    \item We validate the \proposedMappingFramework{} mapping framework in two case studies, volumetric mapping and exploration, and compare it to OctoMap.
\end{itemize}

The remainder of this paper is organized as follows. Related work is presented in Section~\ref{sec:related_work}, while in Section~\ref{sec:mapping_framework} the \proposedMappingFramework{} mapping framework is presented. Implementation details are given in Section~\ref{sec:implementation}, followed by an evaluation of the proposed framework and case studies in Sections~\ref{sec:evaluation} and~\ref{sec:case_studies}, respectively. Finally, Section~\ref{sec:conclusion} summarizes and concludes the paper.

\section{Related Work}
\label{sec:related_work}
This section gives a brief overview of different 3D map representations used in robotics.

A popular approach to model 3D environments is to discretize the world into equal sized cubic volumes \cite{Roth-Tabak1989BuildingInformation}, called voxels. One of the major shortcomings of fixed grid structures is that the size of the area to be mapped has to be known \emph{a-priori}. Memory requirements can also be a problem when mapping large areas at a high resolution. Voxel hashing \cite{Niener2013Real-timeHashing} is one approach to overcome these shortcomings, where fixed sized blocks are allocated on demand.

The mapping framework Voxblox~\cite{Oleynikova2017Voxblox:Planning} uses a signed distance field~\cite{Oleynikova2016SignedPlanning} voxel grid, with voxel hashing for dynamic growth, as representation. It was mainly developed for planning or trajectory optimizations in the context of micro aerial vehicles (MAVs). The signed distance field representation makes trajectory optimizations faster by storing the distance to the closest obstacle in each voxel. Voxblox builds on~\cite{Niener2013Real-timeHashing} where they use a spatial hashing scheme and allocate blocks of fixed size when needed. This means that the size of the area to be mapped does not need to be known beforehand.

One of the most popular mapping frameworks is OctoMap~\cite{hornung2013octomap}. OctoMap uses an octree-based data structure, as proposed in \cite{meagher1982geometric}, to do occupancy mapping. The octree structure allows for delaying the initialization of the grid structure. It is also more memory efficient compared to Voxblox or a fixed grid structure since the information can be stored at different resolutions in the octree, without losing any precision. If the inner nodes of the octree are updated correctly, it is possible to do queries at different resolutions. This can be especially beneficial in systems where multiple algorithms are using the same map but have different computational and time requirements. These are important properties and we will therefore use an octree representation and build on OctoMap.

\section{\proposedMappingFramework{} Mapping Framework}
\label{sec:mapping_framework}
This section presents the \proposedMappingFramework{} mapping framework, which is based on OctoMap. We highlight the difference between the two frameworks to make comparison easier.

\proposedMappingFramework{} uses an octree data structure, just like OctoMap. An octree is a tree data structure where each node represents a cubic volume of the environment being mapped. A node can recursively be split up into eight equal sized smaller nodes, until a minimum size has been reached. The resolution of the octree is the same as the minimum node size. A visual representation of an octree can be seen in Fig.~\ref{fig:ufomap_tree}.

Each node has an occupancy value indicating if the cubic volume represented by the node is occupied, free, or unknown. A probabilistic occupancy value is used in order to better handle sensor noise and dynamic environments. More specifically the log-odds occupancy value is stored in the nodes. Sensor readings can thus be fused in using only additions instead of multiplications. The log-odds occupancy value is clamped to allow for pruning the tree, as in OctoMap. Pruning can be applied when all eight of a node's children are the same. This leads to a smaller tree, which is beneficial both in terms of memory usage and efficiency when traversing the tree.

As mentioned in Section~\ref{sec:related_work}, the octree data structure allows for queries at different resolutions. This can speed-up applications, as well as allow for different applications that require different resolutions to share the same map. As in OctoMap, a node is defined to have the same occupancy value as the maximum occupancy value of its children. Using the maximum value is conservative and it enables doing path planning, or similar, at whichever resolution without missing any obstacles.

The nodes in \proposedMappingFramework{} store three indicators, $i_{f}$, $i_{u}$, and $i_{a}$, which are not found in OctoMap. $i_{f}$ and $i_{u}$ indicates if a node contains free space or unknown space, respectively. No indicator is needed for the occupied space, since the occupancy value of a node is enough to tell if it contains occupied space. The third indicator, $i_{a}$, indicates whether all of a node's children are the same. $i_{a}$ is useful in cases were automatic pruning has been disabled, but you still want the same behaviour as if it were enabled. Automatic pruning means that the octree is automatically pruned when data is deleted or inserted into the tree. The main reason for disabling automatic pruning is for multi-threading, where you want to access and insert/delete data at the same time (see Section~\ref{sec:multi_threading}). It is possible to manually specify when the pruning should occur in this case. The indicators allow for increased efficiency when traversing the tree, by allowing certain branches to be ignored, for example, when only looking for unknown space.

The state of a node, $n$, in OctoMap is determined by a threshold, $t_{o}$:
\[ \text{state}(n) =
  \begin{cases}
    \text{unknown}  & \quad \text{if } n \text{ is null pointer} \\
    \text{occupied} & \quad \text{if } t_{o} \leq \text{occ}(n)  \\
    \text{free}     & \quad \text{else}
  \end{cases}
\]
where $\text{occ}(n)$ is the occupancy value of the node $n$. In \proposedMappingFramework{} this has been expanded to include a threshold, $t_{f}$, for when a node is free as well:
\[ \text{state}(n) =
  \begin{cases}
    \text{unknown}  & \quad \text{if } t_{f} \leq \text{occ}(n) \leq t_{o} \\
    \text{occupied} & \quad \text{if } t_{o} < \text{occ}(n)               \\
    \text{free}     & \quad \text{if } t_{f} > \text{occ}(n)
  \end{cases}
\]
these two thresholds together can be useful in certain applications. In reconstruction or exploration you want to map a certain volume. By setting $t_{f} = 0.1$ and $t_{o} = 0.9$, the nodes stay in the unknown state until they have high confidence that they are either occupied or free. This can simplify the reconstruction or exploration algorithm since only unknown nodes have to be considered. In OctoMap's case, by setting $t_{o} = 0.9$, a discovered node will most likely be classified as free at first. By searching for only unknown nodes in this case the map will likely end up only containing free nodes. The algorithm would therefore have to search through all the free nodes in the map, and usually there are a lot more free nodes than there are occupied nodes.

Alternatively, if you use $t_{f} = 0.49$ and $t_{o} = 0.9$, the exploration algorithm would focus its attention on the occupied space, meaning you would target higher certainty about the occupied space. This can be convenient in applications requiring reconstruction. Note that you do not have to change anything in the exploration method, and instead use these parameters to modify the exploration behaviour.

\begin{figure}
  \centering
  \resizebox{1.0\linewidth}{!}{
    \begin{tikzpicture}[
        node distance=0pt,
        start chain = A going right,
        X/.style = {rectangle, draw,
            minimum width=2ex, minimum height=3ex,
            outer sep=0pt, on chain},
        B/.style = {decorate,
            decoration={brace, amplitude=5pt,
                pre=moveto,pre length=1pt,post=moveto,post length=1pt,
                raise=1mm,
                #1},
            thick},
        B/.default=mirror,
      ]
      \foreach \i in {0, 0,
          $z_{20}$, $y_{20}$, $x_{20}$,
          $z_{19}$, $y_{19}$, $x_{19}$,
          \text{...},
          $z_{1}$, $y_{1}$, $x_{1}$,
          $z_{0}$, $y_{0}$, $x_{0}$}
      \node[X] {\i};
      \draw[B=] ( A-1.north west) -- node[above=2mm] {Unused} ( A-2.north east);
      \draw[B] (A-3.south west) -- node[below=2mm] {Child$_\text{idx}$ depth 20} (A-5.south east);
      \draw[B=] (A-6.north west) -- node[above=2mm] {Child$_\text{idx}$ depth 19} (A-8.north east);
      \draw[B] (A-10.south west) -- node[below=2mm] {Child$_\text{idx}$ depth 1} (A-12.south east);
      \draw[B=] (A-13.north west) -- node[above=2mm] {Child$_\text{idx}$ depth 0} (A-15.north east);
    \end{tikzpicture}
  }
  \caption{Bit string representation of Morton code used in \proposedMappingFramework{}.}
  \label{fig:bit_string}
\end{figure}
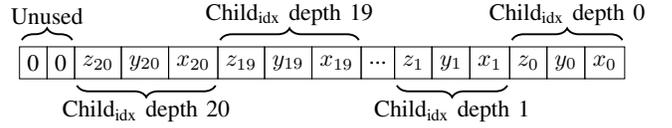

Finally, Morton codes~\cite{Morton1966ASequencing} are used to speedup traversal of the octree. To generate a Morton code, we first convert the cartesian coordinate $c=(c_{x}, c_{y}, c_{z}) \in \Real^{3}$ to the octree node index tuple $k=(k_x,k_y,k_z) \in \Natural_{0}^3$ that $c$ falls into:

\begin{equation}
  \label{eq:coord_to_node_center}
  k_{j} = \floor{\frac{c_{j}}{\text{res}}} + 2^{d_\text{max} - 1}
\end{equation}
where $d_\text{max}$ is the depth of the octree. By interleaving the bits representing $k_{j}$ as shown in Fig.~\ref{fig:bit_string}, the Morton code $m$ is created for $c$. From the root node it is then possible to traverse down to the node by simply looking at the three bits of $m$ that correspond to the depth of the child and moving to that child.

The Morton code generation has been accelerated by using integer dilation and contraction~\cite{Stocco1995IntegerOctrees}.

\section{Implementation Details}
\label{sec:implementation}
This section covers implementation details and highlights differences compared to OctoMap.

\subsection{Nodes}
\label{sec:nodes}
\proposedMappingFramework{} has three different types of nodes; inner nodes, inner leaf nodes, and leaf nodes. In contrast, OctoMap has only what corresponds to the inner nodes and inner leaf nodes.

\subsubsection{Inner nodes}
\label{sec:inner_nodes}
The inner nodes are all the nodes that have children. An inner node contains a 4-byte log-odds occupancy value and an 8-byte pointer to an array of its children. In \proposedMappingFramework{} the children are stored directly in the array, instead of the array holding pointers to the children like in OctoMap. Meaning that 64 bytes are saved for an inner node with 8 children in \proposedMappingFramework{} compared to OctoMap. However, this also means that a node in \proposedMappingFramework{} either must have no children or 8 children.

The inner nodes also contain the three indicators mentioned in Section~\ref{sec:mapping_framework}, taking a total of 4 bytes. On a 64-bit operating system this means that the inner nodes require 16 bytes, compared to 72 bytes for OctoMap.

\subsubsection{Inner leaf nodes}
\label{sec:inner_leaf_nodes}
Inner leaf nodes are exactly the same as the inner nodes, described in Section~\ref{sec:inner_nodes}. The only distinction is that the inner leaf nodes have no children, meaning the child pointer is a null pointer. Once an inner leaf node gets a child it is considered an inner node. In OctoMap this is the only kind of leaf node that exist.

On a 64-bit operating system an inner leaf node in \proposedMappingFramework{} takes up 16 bytes, the same as in OctoMap.

\subsubsection{Leaf nodes}
\label{sec:leaf_nodes}
The leaf nodes are more simplistic. Only the 4-byte log-odds occupancy value is stored. Compared to an inner leaf node, it is not possible for a leaf node to have children. In OctoMap the corresponding node is the inner leaf node, which takes up 16 bytes.

The three kinds of nodes can be seen in Fig.~\ref{fig:ufomap_datastructure}. The nodes which have an occupancy value, indicators, and a child pointer are inner nodes, and where the child pointer is empty is an inner leaf node. The nodes with only an occupancy value are the leaf nodes.

As stated in \cite{hornung2013octomap}, and also seen in Table~\ref{tab:memory_consumption}, 80-85\% of the octree nodes are (inner) leaf nodes in OctoMap. However, since OctoMap does not differentiate between inner leaf nodes and leaf nodes we see that this value drops to 71-81\% in \proposedMappingFramework{}, where the distinction is made and only leaf nodes are shown. This means that the majority of the nodes in the tree are leaf nodes. Therefore, having the dedicated leaf node data structures and by storing the children of an inner node directly in an array in \proposedMappingFramework{} significantly reduces the memory usage over OctoMap by a factor of 3, even though the total amount of nodes in the tree increases (see Table~\ref{tab:memory_consumption}).

\subsection{Multi-threading}
\label{sec:multi_threading}
Multi-core processors are the norm in most segments nowadays. To fully utilize these processors the third indicator $i_{a}$, mentioned in Section~\ref{sec:mapping_framework}, which indicates whether all of an inner node's children are the same, was introduced.

When inserting data into \proposedMappingFramework{} and OctoMap the tree is automatically pruned when possible. If pruning occurs in a thread at the same time as another thread is iterating through the tree, it can remove nodes that the iterator is currently pointing to. Making the iterator invalid, with undefined behavior if dereferenced.

To prevent this, \proposedMappingFramework{} allows for disabling the automatic pruning. In this case $i_{a}$ will be used to determine if a node is an inner node or an inner leaf node. If it is the latter, it will not traverse further down the tree. This allows for inserting data and traversing the tree at the same time, while still utilizing the octree structure. It is possible to manually tell \proposedMappingFramework{} when to prune the tree, such that the memory consumption still can remain low while maintaining thread safety.

\begin{table*}[t]
    \centering
    \caption{Memory consumption and number of nodes comparison between \proposedMappingFramework{} and OctoMap on the OctoMap 3D scan dataset.}
    \begin{tabular}{@{}lccccccccc@{}}
        \toprule

        \multicolumn{1}{l}{\multirow{3}{*}{Dataset}} &  & \multicolumn{3}{c}{\multirow{2}{*}{Memory usage}} &              & \multicolumn{4}{c}{\multirow{2}{*}{Number of nodes}}                                                                                                                  \\
                                                     &  & \multicolumn{3}{c}{}                              &              & \multicolumn{4}{c}{}                                                                                                                                                  \\
        \cmidrule{3-5} \cmidrule{7-10}
                                                     &  & \proposedMappingFramework{} (MB)                  & OctoMap (MB) & Reduction (\%)                                       &  & \proposedMappingFramework{} & OctoMap       & \proposedMappingFramework{} leaves (\%) & OctoMap leaves (\%) \\

        \midrule

        FR-078 tidyup                                &  & \num{7.42}                                        & \num{21.49}  & \num{65.47}                                          &  & \num{1642113}               & \num{1369165} & \num{80.51}                             & \num{85.01}         \\

        FR-079 corridor                              &  & \num{19.70}                                       & \num{51.86}  & \num{62.01}                                          &  & \num{2823713}               & \num{1829134} & \num{75.20}                             & \num{80.70}         \\

        Freiburg campus                              &  & \num{58.71}                                       & \num{155.46} & \num{62.23}                                          &  & \num{8402193}               & \num{5515178} & \num{75.10}                             & \num{80.96}         \\

        freiburg1\_360                               &  & \num{15.98}                                       & \num{42.05}  & \num{62.00}                                          &  & \num{2161849}               & \num{1547112} & \num{71.72}                             & \num{82.53}         \\

        New College                                  &  & \num{29.41}                                       & \num{75.40}  & \num{60.99}                                          &  & \num{4157217}               & \num{2633701} & \num{74.37}                             & \num{80.27}         \\

        \bottomrule
    \end{tabular}
    \label{tab:memory_consumption}
\end{table*}

\subsection{Integrating Point Cloud Sensor Measurements}
\label{sec:integrating_sensor_measurements}
Ray tracing is used to integrate point cloud sensor measurements. When integrating a point cloud it is important that points that should be occupied gets occupied and that all points between the sensor origin and each point of the point cloud gets freed. In \proposedMappingFramework{} there are four methods for this, each faster than the previous but with less accurate results:

\subsubsection{Simple integrator}
\label{sec:simple_integrator}
For each point in the point cloud we trace a ray, using \cite{Amanatides1987ATracing}, and free all nodes from origin to the point. We set the node in which the point lies to occupied. Same as in OctoMap.

\subsubsection{Discrete integrator}
\label{sec:discrete_integrator}
We use the fact that a number of points in the point cloud fall inside the same node. So by discretizing the point cloud first, the ray tracing is only done once for each node the points fall into. The ray tracing is performed towards the center of the node which means the result can be different compared to the simple integration. However, this is significantly faster for large point clouds with many points falling into the same node. The same method exists in OctoMap.

\subsubsection{Fast discrete integrator}
\label{sec:fast_discrete_integrator}
In the fast discrete integrator, the ray tracing and discretization is performed at multiple different depths. First, a ray tracing algorithm (Eq.~\ref{eq:fast_discrete_integrator}), cruder than \cite{Amanatides1987ATracing}, is applied at a coarser resolution, corresponding to depth $d$. Ray tracing is done until $n$ nodes, at that resolution, away from the end node. From that node we perform a new ray tracing at lower depth until we reach depth $0$ (leaf node depth) where we perform the same ray tracing as in the discrete integration. This ensures that we do not clear space behind the occupied space. The parameters $n$ and $d$ allow us to trade speed against accuracy. 
\begin{equation}
    \label{eq:fast_discrete_integrator}
    \begin{split}
        \text{free}_{i} = \text{origin} + i \cdot \text{res}_{d} \cdot \frac{\left( \text{end} - \text{origin} \right)}{\left\lVert \text{end} - \text{origin} \right\rVert}, \\ \quad \forall i \in {0,\dots,\frac{\left\lVert \text{end} - \text{origin} \right\rVert}{\text{res}_{d}} - n}
    \end{split}
\end{equation}
where $\text{res}_{d}$ is the resolution at the depth $d$. The special case $d=0$ corresponds to the discrete integrator above and in the case $n=0$ space is cleared only at depth $d$.

Extra care has to be taken when inserting data into the octree at a different depth, as in the fast discrete integrator method. When clearing free space at depth $d$ we check if the node at that depth is not occupied. If it is not occupied, we can simply update the occupancy value of that node, and remove its children. If it is occupied, we will instead go down one more depth level and perform the same check until we are at a leaf node.

The occupied nodes are updated at the highest resolution in all three integrators. Thus, the occupied space will look mostly the same for all three methods, which is the most important aspect in most applications.

\begin{figure}
    \centering
    \subfloat[]{
        \includegraphics[angle=-90,width=0.3\linewidth]{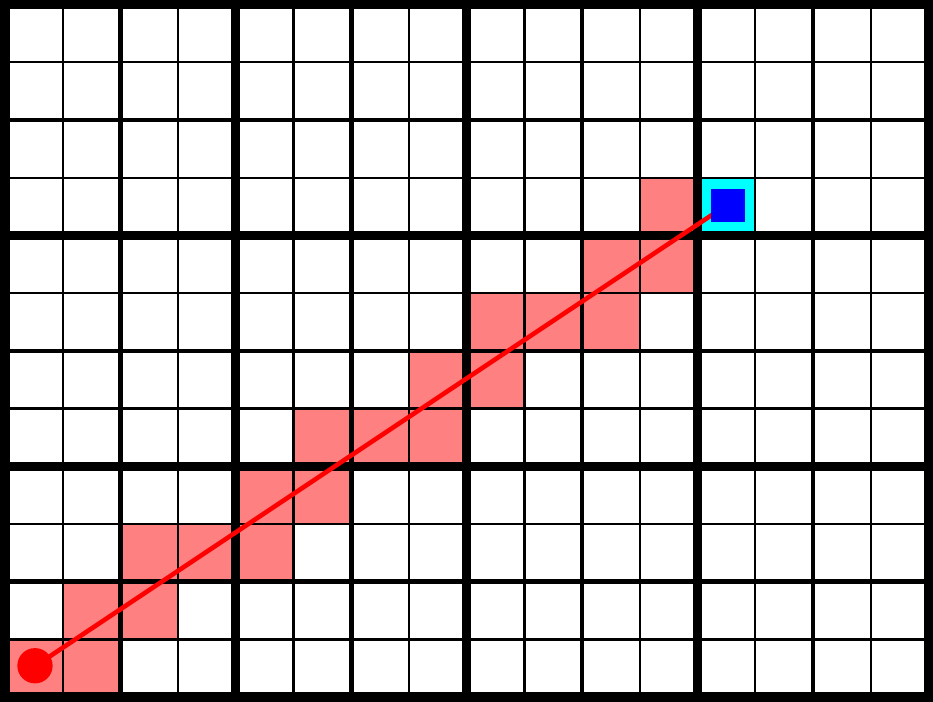}
        \label{fig:integration_comparision_simple}
    }
    \subfloat[]{
        \includegraphics[angle=-90,width=0.3\linewidth]{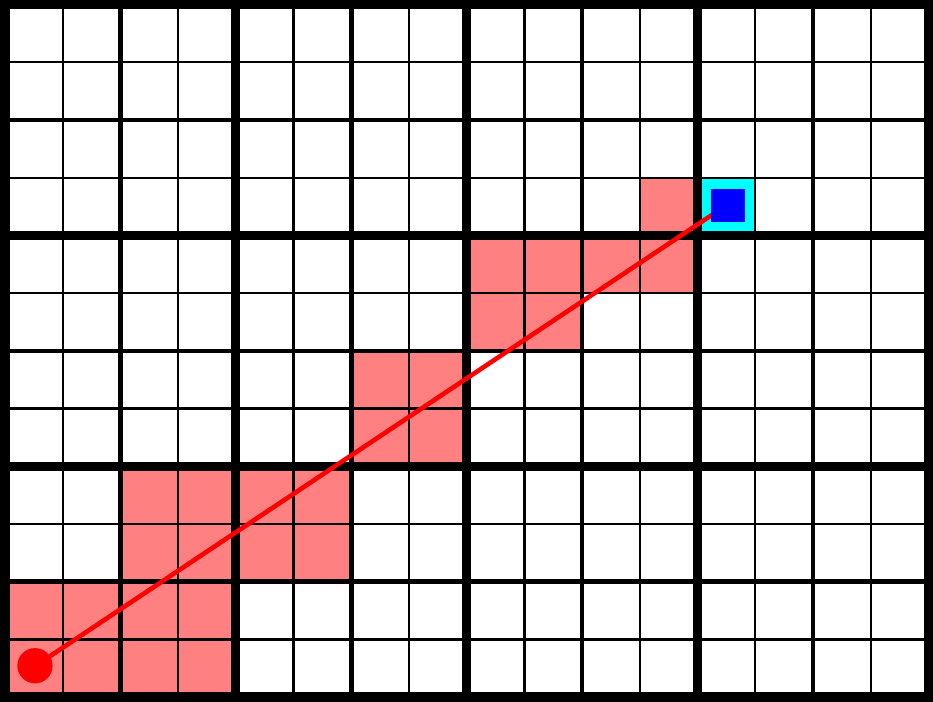}
        \label{fig:integration_comparision_fast}
    }
    \subfloat[]{
        \includegraphics[angle=-90,width=0.3\linewidth]{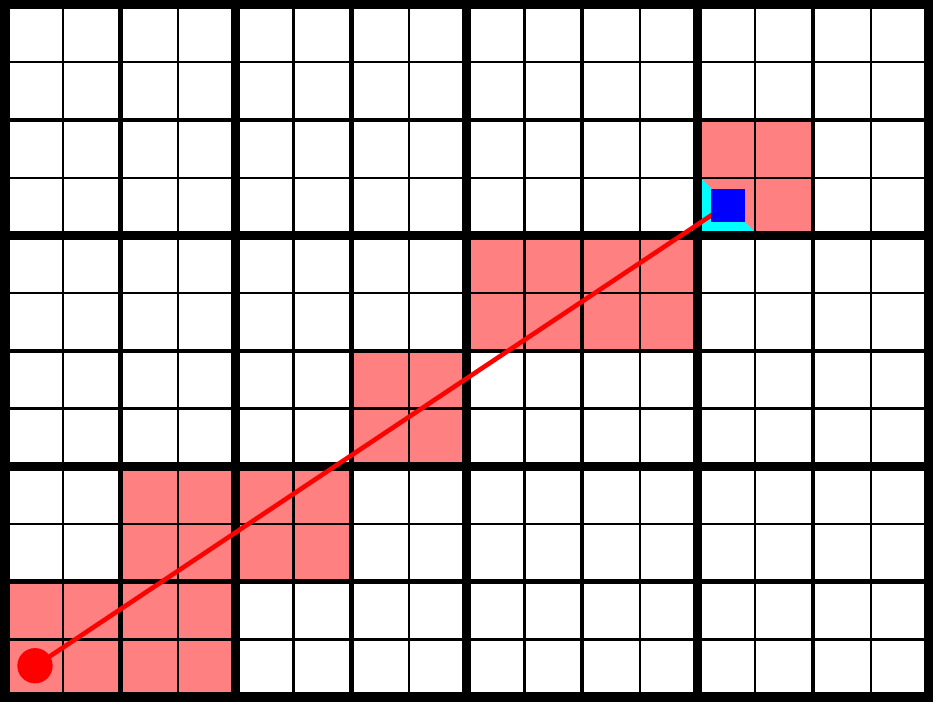}
        \label{fig:integration_comparision_super}
    }
    \caption{Comparison of the integrators mentioned in Section~\ref{sec:integrating_sensor_measurements}, for a single ray starting at the bottom left corner (red circle) and ending near the top right corner (blue square). The red line is the line being traced. The red and blue cell(s) will be marked free or occupied by the integrator, respectively. \protect\subref{fig:integration_comparision_simple} Simple/discrete integrator \protect\subref{fig:integration_comparision_fast} Fast discrete integrator with $d=1$ and $n=2$ \protect\subref{fig:integration_comparision_super} Fast discrete integrator with $d=1$ and $n=0$. Note that the end point will be marked both as free and occupied.}
    \label{fig:integration_comparision}
\end{figure}

A comparison of the integrators can be seen in Fig.~\ref{fig:integration_comparision}.

\subsection{Bounding Box}
\label{sec:bounding_box}
\proposedMappingFramework{} and OctoMap supports defining a bounding box to speedup operations. Only a subset of the octree has to be searched and updated. When inserting data into OctoMap only rays that begin and end inside the bounding box are integrated into the map. This can be a problem for exploration algorithms which rely on free space being cleared. In \proposedMappingFramework{}, this problem is fixed, by moving the beginning and ending of the rays towards each other until they both are inside the bounding box.

\subsection{Accessing Data}
\label{sec:accessing_data}
To access the data in \proposedMappingFramework{}, iterators are used. They are fast and they can be used to specify bounding boxes, bounding spheres and bounding frustums. Only bounding boxes are available in OctoMap. Iterators can also specify the type of nodes that should be returned, e.g., only leaves, occupied, free, unknown, contains occupied/free/unknown, or a combination of them. In comparison, OctoMap always returns free and occupied nodes. It is not possible to get the unknown nodes, and it is not possible to get only occupied nodes or only free nodes.

\subsection{Availability}
\label{sec:availability}
\proposedMappingFramework{} is available as a self-contained C++ library at \url{https://github.com/danielduberg/UFOMap}. Packages for integration with the Robot Operation System (ROS) \cite{ros} are also available. There are functions for reading and writing OctoMap files and converting between \proposedMappingFramework{} and OctoMap. This facilitates the transition from OctoMap to \proposedMappingFramework{} in an already existing system, and allows to utilize both mapping frameworks for different parts of the same system.

\section{Evaluation}
\label{sec:evaluation}
We compare our proposed mapping framework, \proposedMappingFramework{}, against OctoMap when it comes to memory consumption, insertion time, and in three different use cases.

\subsection{Memory Consumption and Node Count}
\label{sec:memory_consumption_and_node_count}
In the first experiment, we compare the memory consumption between \proposedMappingFramework{} and OctoMap on the OctoMap 3D scan dataset\footnote{\url{http://ais.informatik.uni-freiburg.de/projects/datasets/octomap}}. We analyze the memory usage when the tree has been pruned. Table~\ref{tab:memory_consumption} shows that \proposedMappingFramework{} is around 38\% of the size of OctoMap, even though UFOMap contains more nodes in total, as seen in the same table. The increase in the number of nodes is a result of the unknown space being explicitly represented, meaning an inner node must have either 0 or 8 children. The decrease in memory usage is because of i) the smaller data structures for the leaf nodes mentioned in Section~\ref{sec:nodes} and ii) storing the child nodes directly in the child array of the inner nodes, instead of storing pointers to the children as in OctoMap.

\begin{table}[H]
    \centering
    \caption{Insertion timings comparison between \proposedMappingFramework{} and OctoMap, on the cow dataset$^\text{\protect\ref{fn:cow_dataset}}$, with different voxel sizes.}
    \begin{tabular}{@{}l@{\hspace{0.3\tabcolsep}}cccc@{}}
        \toprule
        
        Method & \thead{Voxel size\\(\SI{}{\centi\metre})} & \thead{Total\\(\SI{}{\milli\second})} & \thead{Ray tracing\\(\SI{}{\milli\second})} & \thead{Insertion\\(\SI{}{\milli\second})} \\
        
        \midrule
        
        \proposedMappingFramework{} & \multirow{2}{*}{16} & 
        \num[round-mode=places,round-precision=2]{4.9816896908}$\pm$\num[round-mode=places,round-precision=2]{1.2985754560} & 
        \num[round-mode=places,round-precision=2]{4.6403627053}$\pm$\num[round-mode=places,round-precision=2]{1.0804954384} & 
        \num[round-mode=places,round-precision=2]{0.3413269855}$\pm$\num[round-mode=places,round-precision=2]{0.2660504570} \\

        OctoMap & & 
        \num[round-mode=places,round-precision=2]{5.5182577406}$\pm$\num[round-mode=places,round-precision=2]{1.6576842461} & 
        \num[round-mode=places,round-precision=2]{4.8608115864}$\pm$\num[round-mode=places,round-precision=2]{1.1840283685} & 
        \num[round-mode=places,round-precision=2]{0.6574461542}$\pm$\num[round-mode=places,round-precision=2]{0.5583532011} \vspace{0.15cm}\\
        
        \proposedMappingFramework{} & \multirow{4}{*}{8} & 
        \num[round-mode=places,round-precision=1]{12.3235210338}$\pm$\num[round-mode=places,round-precision=1]{7.4157542385} & 
        \num[round-mode=places,round-precision=1]{10.4342135277}$\pm$\num[round-mode=places,round-precision=1]{5.8203050528} & 
        \num[round-mode=places,round-precision=1]{1.8893075061}$\pm$\num[round-mode=places,round-precision=1]{1.6384820888} \\

        OctoMap & & 
        \num[round-mode=places,round-precision=1]{16.2654127707}$\pm$\num[round-mode=places,round-precision=1]{10.4309395519} & 
        \num[round-mode=places,round-precision=1]{12.1584668952}$\pm$\num[round-mode=places,round-precision=1]{6.8510765204} & 
        \num[round-mode=places,round-precision=1]{4.1069458755}$\pm$\num[round-mode=places,round-precision=1]{3.7045717100} \\

        \proposedMappingFramework{}$^\star$ & & 
        \num[round-mode=places,round-precision=1]{8.1203142493}$\pm$\num[round-mode=places,round-precision=1]{3.3968101671} & 
        \num[round-mode=places,round-precision=1]{6.6827449238}$\pm$\num[round-mode=places,round-precision=1]{2.2989153317} & 
        \num[round-mode=places,round-precision=1]{1.4375693255}$\pm$\num[round-mode=places,round-precision=1]{1.1477244475} \\

        \proposedMappingFramework{}$^\dagger$ & & 
        \num[round-mode=places,round-precision=1]{6.4855423690}$\pm$\num[round-mode=places,round-precision=1]{2.1933600364} & 
        \num[round-mode=places,round-precision=1]{5.9524178618}$\pm$\num[round-mode=places,round-precision=1]{1.8205885492} & 
        \num[round-mode=places,round-precision=1]{0.5331245072}$\pm$\num[round-mode=places,round-precision=1]{0.3985462977} \vspace{0.15cm}\\
        
        \proposedMappingFramework{} & \multirow{4}{*}{4} &  
        \num[round-mode=places,round-precision=1]{60.9331559127}$\pm$\num[round-mode=places,round-precision=1]{44.6906539764} & 
        \num[round-mode=places,round-precision=1]{46.7966853493}$\pm$\num[round-mode=places,round-precision=1]{32.7431565070} & 
        \num[round-mode=places,round-precision=1]{14.1364705634}$\pm$\num[round-mode=places,round-precision=1]{12.1824597679} \\

        OctoMap & & 
        \num[round-mode=places,round-precision=1]{104.6367209049}$\pm$\num[round-mode=places,round-precision=1]{82.1661080323} & 
        \num[round-mode=places,round-precision=1]{71.7608723846}$\pm$\num[round-mode=places,round-precision=1]{53.0442499679} & 
        \num[round-mode=places,round-precision=1]{32.8758485203}$\pm$\num[round-mode=places,round-precision=1]{30.0643343850} \\

        \proposedMappingFramework{}$^\star$ & &  
        \num[round-mode=places,round-precision=1]{21.1068514954}$\pm$\num[round-mode=places,round-precision=1]{11.9671644413} & 
        \num[round-mode=places,round-precision=1]{14.2729841342}$\pm$\num[round-mode=places,round-precision=1]{6.6977463377} & 
        \num[round-mode=places,round-precision=1]{6.8338673612}$\pm$\num[round-mode=places,round-precision=1]{5.3529957065} \\

        \proposedMappingFramework{}$^\dagger$ & &  
        \num[round-mode=places,round-precision=1]{10.9259971825}$\pm$\num[round-mode=places,round-precision=1]{4.7014055498} & 
        \num[round-mode=places,round-precision=1]{9.2931545355}$\pm$\num[round-mode=places,round-precision=1]{3.4898880952} & 
        \num[round-mode=places,round-precision=1]{1.6328426470}$\pm$\num[round-mode=places,round-precision=1]{1.2774703456} \vspace{0.15cm}\\
        
        \proposedMappingFramework{} & \multirow{4}{*}{2} & 
        \num[round-mode=places,round-precision=0]{371.0695696013}$\pm$\num[round-mode=places,round-precision=0]{254.0529319326} & 
        \num[round-mode=places,round-precision=0]{263.9060054266}$\pm$\num[round-mode=places,round-precision=0]{176.4307166460} & 
        \num[round-mode=places,round-precision=0]{107.1635641747}$\pm$\num[round-mode=places,round-precision=0]{79.4085384163} \\

        OctoMap & & 
        \num[round-mode=places,round-precision=0]{745.2208940539}$\pm$\num[round-mode=places,round-precision=0]{548.0002262905} & 
        \num[round-mode=places,round-precision=0]{520.7813661572}$\pm$\num[round-mode=places,round-precision=0]{368.7314573804} & 
        \num[round-mode=places,round-precision=0]{224.4395278967}$\pm$\num[round-mode=places,round-precision=0]{188.4295372348} \\

        \proposedMappingFramework{}$^\star$ & & 
        \num[round-mode=places,round-precision=0]{74.0010506878}$\pm$\num[round-mode=places,round-precision=0]{43.8805164934} & 
        \num[round-mode=places,round-precision=0]{42.1458407243}$\pm$\num[round-mode=places,round-precision=0]{22.3444462352} & 
        \num[round-mode=places,round-precision=0]{31.8552099636}$\pm$\num[round-mode=places,round-precision=0]{21.9006177227} \\

        \proposedMappingFramework{}$^\dagger$ & & 
        \num[round-mode=places,round-precision=0]{28.4052115043}$\pm$\num[round-mode=places,round-precision=0]{15.2185052571} & 
        \num[round-mode=places,round-precision=0]{19.6875502059}$\pm$\num[round-mode=places,round-precision=0]{8.7865373155} & 
        \num[round-mode=places,round-precision=0]{8.7176612984}$\pm$\num[round-mode=places,round-precision=0]{6.7080721545} \\

        \bottomrule
    \end{tabular}
    \label{tab:insertion}
\end{table}

\FloatBarrier

\subsection{Insertion}
\label{sec:insertion}
Point cloud integration time comparison between \proposedMappingFramework{} and OctoMap is shown in Table~\ref{tab:insertion}, using the cow dataset\footnote{\label{fn:cow_dataset}\url{https://projects.asl.ethz.ch/datasets/doku.php?id=iros2017}}. For both \proposedMappingFramework{} and OctoMap the discrete integrator, mentioned in Section~\ref{sec:discrete_integrator}, is used. \proposedMappingFramework{}$^\star$ uses the fast discrete integrator, mentioned in Section~\ref{sec:fast_discrete_integrator}, with $n=2$ and $d$ corresponding to the depth of the voxels at \SI{16}{\centi\metre} voxel size. \proposedMappingFramework{}$^\dagger$ uses the fast integrator with $n=0$ and again $d$ corresponding to the depth of the voxels at \SI{16}{\centi\metre} voxel size.

The total time represents the average time for a single point cloud to be integrated. Ray tracing shows the average time per point cloud for doing the ray tracing part of the insertion. Insertion shows the average time per point cloud for integrating the points calculated from the ray tracing step into the octree. The standard deviation is included for all.

When only looking at the discrete integrator for \proposedMappingFramework{}, we see that it is about two times faster at the insertion part of the integration compared to OctoMap. The ray tracing is between 1 to 2 times faster in \proposedMappingFramework{}, which is, most likely, due to different implementation factors since both utilize the same algorithm. As the voxel size decreases, the difference between the two framework increases, in favour of \proposedMappingFramework{}. At a voxel size of \SI{4}{\centi\metre} and below, the need for the faster integrators becomes very apparent.

\proposedMappingFramework{}$^\star$ and \proposedMappingFramework{}$^\dagger$ provides just that, fast integration while still clearing free space. With \proposedMappingFramework{}$^\dagger$ scaling a lot better with the resolution and being 26 times faster than OctoMap at \SI{2}{\centi\metre} voxel size.

\subsection{Collision Checking}
\label{sec:check_robot_area}
In the first of the use cases we check if the robot, or a part of the robot, can be at a certain position in the map. That is, we check if a region of the map is free, meaning there is no occupied or unknown space in that region. This is an operation that is heavily used in sampling-based motion planners, such as RRTs. By not allowing any unknown space in this region we are more conservative and safe. We sampled \num{1000000} poses where at least the center of the pose was in free space. A radius of \SI{25}{\centi\metre} was used. On average around 50\% of the poses sampled were in collision. The results are presented in Table~\ref{tab:collision_checking}. We can see that \proposedMappingFramework{} allows for faster collision checking than OctoMap in all cases but the ones with large voxels. This is most likely because of the overhead for constructing the iterators in \proposedMappingFramework{} in these cases. The last column shows the result when using the same way to traverse the octree for collisions as in OctoMap in \proposedMappingFramework{}. We see that this is faster than OctoMap for all resolutions. However, it is significantly slower than the default method in \proposedMappingFramework{}, which exploits the octree structure, for the higher resolutions.

The reason for \proposedMappingFramework{} seemingly being invariant to the voxel size can be because of the indicators $i_{f}$ and $i_{u}$, presented in Section~\ref{sec:mapping_framework}, together with the iterators mentioned in Section~\ref{sec:accessing_data}. By specifying the bounding box and that only occupied and unknown voxels should be retrieved, \proposedMappingFramework{} can move straight down the octree to a node that is either occupied or unknown inside the radius.

\begin{table}[t]
    \centering
    \caption{Comparison of time taken to do collision checking in \proposedMappingFramework{} compared to OctoMap.}
    \begin{tabular}{@{}l@{\hspace{0.3\tabcolsep}}crrr@{}}
        \toprule

        Dataset & \thead{Voxel size\\(\SI{}{\centi\metre})} & \thead{\proposedMappingFramework{}\\(\SI{}{\micro\second}/pose)} & \thead{OctoMap\\(\SI{}{\micro\second}/pose)} & \thead{\proposedMappingFramework{}$^\text{octo}$\\(\SI{}{\micro\second}/pose)} \\

        \midrule

        FR-078 tidyup & 5 &
        \num[round-mode=places,round-precision=1,math-rm=\mathbf]{2.7206297030} & \num[round-mode=places,round-precision=1]{23.0610030760} & 
        \num[round-mode=places,round-precision=1]{14.5078994390} \\
        
        FR-079 corridor & 5 &
        \num[round-mode=places,round-precision=1,math-rm=\mathbf]{2.9802373660} & 
        \num[round-mode=places,round-precision=1]{15.7317617500} & 
        \num[round-mode=places,round-precision=1]{10.3637455370} \\
        
        Freiburg campus & 20 &
        \num[round-mode=places,round-precision=1]{2.3987754440} & 
        \num[round-mode=places,round-precision=1]{1.4424044490} & 
        \num[round-mode=places,round-precision=1,math-rm=\mathbf]{0.7671919790} \\
        
        freiburg1\_360 & 2 &
        \num[round-mode=places,round-precision=1,math-rm=\mathbf]{3.0650641090} & 
        \num[round-mode=places,round-precision=1]{163.2705933250} & 
        \num[round-mode=places,round-precision=1]{121.8060600520} \\
        
        New College & 20 &
        \num[round-mode=places,round-precision=1]{2.5356196620} & 
        \num[round-mode=places,round-precision=1]{1.3956904760} & 
        \num[round-mode=places,round-precision=1,math-rm=\mathbf]{0.7596716100} \\
        \bottomrule
    \end{tabular}
    \label{tab:collision_checking}
\end{table}

For comparison reasons we have also included the time taken if only checking a region for occupied space. This corresponds to a simplification often made in collision checking to speed up the computations. The results are presented in Table~\ref{tab:collision_checking_occupied_only}. We can see that \proposedMappingFramework{} is faster than OctoMap at all resolutions. Reasons for this can be the use of Morton codes for traversing the octree, perhaps less cache misses since less memory is used, or the inclusion of the indicators. Both \proposedMappingFramework{} and OctoMap exploit the octree structure in this experiment. Hence, there is no need for \proposedMappingFramework{}$^{\text{octo}}$.

\begin{table}[t]
    \centering
    \caption{Same as Table~\ref{tab:collision_checking} except that the collision checking is only done w.r.t. occupied space.}
    \begin{tabular}{@{}l@{\hspace{0.3\tabcolsep}}crr@{}}
        \toprule

        Dataset & \thead{Voxel size\\(\SI{}{\centi\metre})} & \thead{\proposedMappingFramework{}\\(\SI{}{\micro\second}/pose)} & \thead{OctoMap\\(\SI{}{\micro\second}/pose)} \\

        \midrule

        FR-078 tidyup & 5 &
        \num[round-mode=places,round-precision=1,math-rm=\mathbf]{1.6496748390} & \num[round-mode=places,round-precision=1]{2.2974412540} \\
        
        FR-079 corridor & 5 &
        \num[round-mode=places,round-precision=1,math-rm=\mathbf]{2.0828529670} & 
        \num[round-mode=places,round-precision=1]{3.6129647930} \\
        
        Freiburg campus & 20 &
        \num[round-mode=places,round-precision=1,math-rm=\mathbf]{1.3694412050} & 
        \num[round-mode=places,round-precision=1]{1.5905016360} \\
        
        freiburg1\_360 & 2 &
        \num[round-mode=places,round-precision=1,math-rm=\mathbf]{2.1949482970} & 
        \num[round-mode=places,round-precision=1]{5.3037641140} \\
        
        New College & 20 &
        \num[round-mode=places,round-precision=1,math-rm=\mathbf]{1.3764187710} & 
        \num[round-mode=places,round-precision=1]{1.5180857730} \\

        \bottomrule
    \end{tabular}
    \label{tab:collision_checking_occupied_only}
\end{table}

\subsection{Collision Checking Along a Line Segment}
\label{sec:path_planning}
In the second use case we perform collision checking along a line. For simple RRT path planning, the operations described in Sections~\ref{sec:check_robot_area} and~\ref{sec:path_planning} are combined.

As in Section~\ref{sec:check_robot_area}, we are conservative and require that there is no occupied or unknown space along the line. The line is defined by two randomly sampled points. As seen in Table~\ref{tab:path_planning} \proposedMappingFramework{} is between 2 to 15 times faster than OctoMap. 

\begin{table}[h]
    \centering
    \caption{Comparison of time taken to do collision checking along a line in \proposedMappingFramework{} compared to OctoMap.}
    \begin{tabular}{@{}l@{\hspace{0.1\tabcolsep}}c@{\hspace{0.1\tabcolsep}}rrr@{}}
        \toprule

        Dataset & \thead{Voxel size\\(\SI{}{\centi\metre})} & \thead{\proposedMappingFramework{}\\(\SI{}{\micro\second}/line)} & \thead{OctoMap\\(\SI{}{\micro\second}/line)} & \thead{\proposedMappingFramework{}$^\text{octo}$\\(\SI{}{\micro\second}/line)} \\

        \midrule

        FR-078 tidyup & 5 &
        \num[round-mode=places,round-precision=0,math-rm=\mathbf]{95.8455363000}$\pm$\num[round-mode=places,round-precision=0,math-rm=\mathbf]{213.8949746840} & 
        \num[round-mode=places,round-precision=0]{845.4210066000}$\pm$\num[round-mode=places,round-precision=0]{1956.0680736361} & 
        \num[round-mode=places,round-precision=0]{568.5921173000}$\pm$\num[round-mode=places,round-precision=0]{1315.1857955210} \\
        
        FR-079 corridor & 5 &
        \num[round-mode=places,round-precision=0,math-rm=\mathbf]{266.7084299000}$\pm$\num[round-mode=places,round-precision=0,math-rm=\mathbf]{686.4590500593} & 
        \num[round-mode=places,round-precision=0]{669.4865106000}$\pm$\num[round-mode=places,round-precision=0]{1844.7180867480} & 
        \num[round-mode=places,round-precision=0]{434.0065101000}$\pm$\num[round-mode=places,round-precision=0]{1186.7686776873} \\
        
        Freiburg campus & 20 &
        \num[round-mode=places,round-precision=0,math-rm=\mathbf]{1306.9673345000}$\pm$\num[round-mode=places,round-precision=0,math-rm=\mathbf]{2476.0131027498} & 
        \num[round-mode=places,round-precision=0]{2576.8262458000}$\pm$\num[round-mode=places,round-precision=0]{5432.1035858416} & 
        \num[round-mode=places,round-precision=0]{1706.8714240000}$\pm$\num[round-mode=places,round-precision=0]{3604.3289797647} \\
        
        freiburg1\_360 & 2 &
        \num[round-mode=places,round-precision=0,math-rm=\mathbf]{112.1768583000}$\pm$\num[round-mode=places,round-precision=0,math-rm=\mathbf]{286.0519258324} & 
        \num[round-mode=places,round-precision=0]{1674.6691251000}$\pm$\num[round-mode=places,round-precision=0]{4300.0152714352} & 
        \num[round-mode=places,round-precision=0]{1134.2616425000}$\pm$\num[round-mode=places,round-precision=0]{2926.0845080292} \\
        
        New College & 20 &
        \num[round-mode=places,round-precision=0,math-rm=\mathbf]{429.7164714000}$\pm$\num[round-mode=places,round-precision=0,math-rm=\mathbf]{824.1373057872} & 
        \num[round-mode=places,round-precision=0]{3526.1817414000}$\pm$\num[round-mode=places,round-precision=0]{6021.3693012379} & 
        \num[round-mode=places,round-precision=0]{2410.2299600000}$\pm$\num[round-mode=places,round-precision=0]{4155.3430107395} \\

        \bottomrule
    \end{tabular}
    \label{tab:path_planning}
\end{table}

As a final comparison, we look at some extreme cases. If the map is completely unknown with a voxel size of \SI{5}{\centi\metre}, it takes \proposedMappingFramework{} \num[round-mode=places,round-precision=2]{0.1756582000} $\SI{}{\micro\second}/\text{line}$, compared with \num[round-mode=places,round-precision=2]{99.4890281000} $\SI{}{\micro\second}/\text{line}$ for OctoMap, and \num[round-mode=places,round-precision=2]{101.5073122000} $\SI{}{\micro\second}/\text{line}$ with \proposedMappingFramework{}$^\text{octo}$. When the map is full of free space it takes \num[round-mode=places,round-precision=0]{34.6900192000} $\SI{}{\micro\second}/\text{pose}$ for \proposedMappingFramework{}, \num[round-mode=places,round-precision=0]{45136.4890129998} $\SI{}{\micro\second}/\text{line}$ for OctoMap, and \num[round-mode=places,round-precision=0]{32491.9270394001} $\SI{}{\micro\second}/\text{line}$ for \proposedMappingFramework{}$^\text{octo}$. In the less likely case when the map is full of occupied space it takes \num[round-mode=places,round-precision=0]{26.4982971000} $\SI{}{\micro\second}/\text{line}$ for \proposedMappingFramework{}, \num[round-mode=places,round-precision=0]{230.7598219000} $\SI{}{\micro\second}/\text{line}$ for OctoMap, and \num[round-mode=places,round-precision=0]{158.4899923000} $\SI{}{\micro\second}/\text{line}$ for \proposedMappingFramework{}$^\text{octo}$.

\begin{table*}[t]
    \centering
    \caption{Comparison of time taken to compute information gain at a pose in \proposedMappingFramework{} compared to OctoMap.}
    \begin{tabular}{@{}l@{\hspace{0.1\tabcolsep}}crrrr@{}}
        \toprule
        Dataset & \thead{Voxel size\\(\SI{}{\centi\metre})} & \thead{\proposedMappingFramework{}\\(\SI{}{\second})} & \thead{\proposedMappingFramework{}$^\text{fast}$\\(\SI{}{\second})} & \thead{OctoMap\\(\SI{}{\second})} & \thead{\proposedMappingFramework{}$^\text{octo}$\\(\SI{}{\second})}\\

        \midrule

        FR-078 tidyup & 5 &
        \num[round-mode=places,round-precision=3]{1.5083057571}$\pm$\num[round-mode=places,round-precision=3]{0.8518477183} & 
        \num[round-mode=places,round-precision=3]{0.6765012937}$\pm$\num[round-mode=places,round-precision=3]{0.3354935130} & 
        \num[round-mode=places,round-precision=3]{5.3863402279}$\pm$\num[round-mode=places,round-precision=3]{0.8531876887} & 
        \num[round-mode=places,round-precision=3]{3.2526645005}$\pm$\num[round-mode=places,round-precision=3]{0.4969856683}\\
        
        FR-079 corridor & 5 &
        \num[round-mode=places,round-precision=3]{1.7050555659}$\pm$\num[round-mode=places,round-precision=3]{0.8216849830} & 
        \num[round-mode=places,round-precision=3]{0.7512161058}$\pm$\num[round-mode=places,round-precision=3]{0.1723491767} & 
        \num[round-mode=places,round-precision=3]{5.3716166660}$\pm$\num[round-mode=places,round-precision=3]{1.3668742319} & 
        \num[round-mode=places,round-precision=3]{3.2426971275}$\pm$\num[round-mode=places,round-precision=3]{0.7909143676} \\
       
        Freiburg campus & 20 &
        \num[round-mode=places,round-precision=3]{0.0079484918}$\pm$\num[round-mode=places,round-precision=3]{0.0045830898} & 
        \num[round-mode=places,round-precision=3]{0.0070711700}$\pm$\num[round-mode=places,round-precision=3]{0.0014712671} & 
        \num[round-mode=places,round-precision=3]{0.0378582099}$\pm$\num[round-mode=places,round-precision=3]{0.0046854702} & 
        \num[round-mode=places,round-precision=3]{0.0247900817}$\pm$\num[round-mode=places,round-precision=3]{0.0030895812} \\
        
        freiburg1\_360 & 2 &
        \num[round-mode=places,round-precision=3]{47.1572934230}$\pm$\num[round-mode=places,round-precision=3]{44.3106318537} & 
        \num[round-mode=places,round-precision=3]{12.4357534542}$\pm$\num[round-mode=places,round-precision=3]{12.8485121774} & 
        \num[round-mode=places,round-precision=3]{183.5542247868}$\pm$\num[round-mode=places,round-precision=3]{65.3765060226} & 
        \num[round-mode=places,round-precision=3]{105.0309074532}$\pm$\num[round-mode=places,round-precision=3]{36.0104512546} \\
        
        New College & 20 &
        \num[round-mode=places,round-precision=3]{0.0081096530}$\pm$\num[round-mode=places,round-precision=3]{0.0038716184} & 
        \num[round-mode=places,round-precision=3]{0.0087846475}$\pm$\num[round-mode=places,round-precision=3]{0.0038006989} & 
        \num[round-mode=places,round-precision=3]{0.0355703553}$\pm$\num[round-mode=places,round-precision=3]{0.0035023238} & 
        \num[round-mode=places,round-precision=3]{0.0238273548}$\pm$\num[round-mode=places,round-precision=3]{0.0020130421} \\

        \bottomrule
    \end{tabular}
    \label{tab:information_gain}
\end{table*}

\subsection{Calculate Information Gain}
\label{sec:calculate_information_gain}
In reconstruction and exploration applications, next-best-view~\cite{connolly1985determination} planning is a popular approach. The next-best-view is often obtained by calculating the information gain from being at a specific pose in the map. The information gain is a measure of how much new information can be collected from a certain pose. In exploration, where the goal is to turn each voxel into either occupied or free space, the information gain can simply be how many unknown nodes can be seen from a certain pose. In this experiment we compare the performance when calculating the information gain in \proposedMappingFramework{} compared to OctoMap. For the sensor we use a horizontal field of view of 115$^\text{o}$ and vertical field of view of 60$^\text{o}$. The minimum and maximum range of the sensor were set to \SI{0}{\metre} and \SI{6.5}{\metre}, respectively.

The results from this experiment can be seen in Table~\ref{tab:information_gain}. OctoMap and \proposedMappingFramework{}$^{\text{octo}}$ compute the information gain similar to how it is done in \cite{bircher2016receding}. Both not exploiting the octree structure.

For \proposedMappingFramework{} we exploit the octree structure to quickly find the unknown voxels inside the region of interest. For each node found we do ray tracing from the sensor to the node, at the same depth in the octree as the node. If the node is not blocked by any occupied space we add it to the total gain, otherwise we recurs down to the node's children and do the ray tracing for each child at their depth instead. We do this until the node is either not blocked by an occupied node or until we are at the leaf depth.

\proposedMappingFramework{}$^{\text{fast}}$ also exploits the octree structure. For each unknown node found, we recurs down to the leaf nodes right away and do the ray tracing. Once a single of the children is not blocked we assume that we can see all children and add all of them to the total gain. Therefore, this approach gives more of an approximate answer than the other.

As seen in Table~\ref{tab:information_gain} the information gain computation is 1.5 to 15 faster with \proposedMappingFramework{} compared to OctoMap, depending on which \proposedMappingFramework{} method is used.

\section{Case Studies}
\label{sec:case_studies}
In this section we compare \proposedMappingFramework{} and OctoMap on two larger use cases.

\begin{figure}[b]
  \centering
  \includegraphics[width=\linewidth]{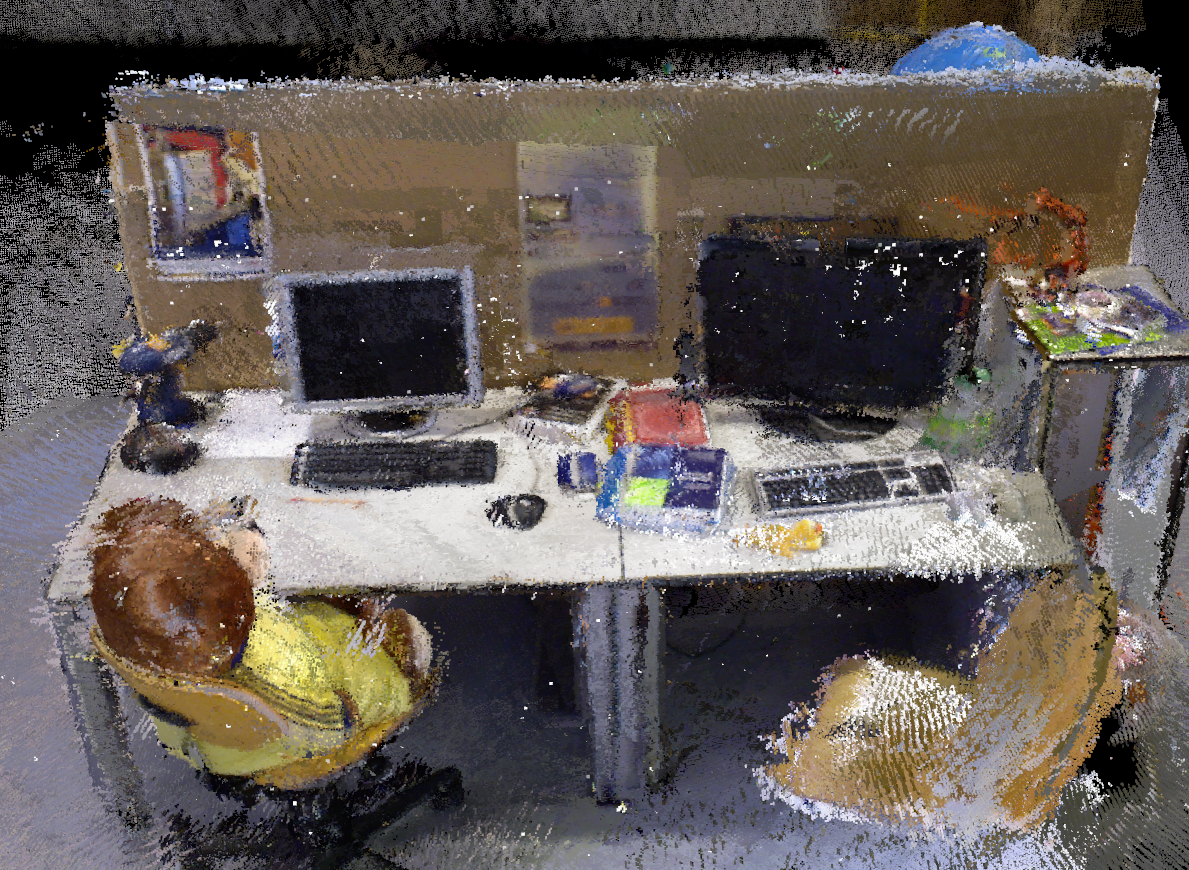}
  \caption{\proposedMappingFramework{} volumetric color mapping at \SI{4}{\milli\metre} voxel size.}
  \label{fig:real_time_mapping}
\end{figure}

\subsection{Real-time Volumetric Mapping}
\label{sec:real_time_volumetric_mapping}
In the first test, we compare the performance on volumetric mapping on the freiburg3\_long\_office\_household sequence from \cite{Sturm2012ASystems}. It is a \SI{86}{\second} office sequence with point cloud data at \SI{2}{\hertz}.

\proposedMappingFramework{} managed to incorporate all point clouds and create a 3D map with \SI{4}{\milli\metre} voxel size with color information at \SI{2}{\hertz}, seen in Fig.~\ref{fig:real_time_mapping}. Without color, \SI{2}{\milli\metre} voxel size without missing any of the point clouds were the limit for \proposedMappingFramework{}. In both cases the fast discrete integrator were used, mentioned in Section~\ref{sec:fast_discrete_integrator}, with $n=0$ and $d=4$. OctoMap were able to run \SI{3}{\centi\metre}, without color, without missing any point cloud. At \SI{2}{\centi\metre} OctoMap could not keep up with the point clouds.

\proposedMappingFramework{} allows specifying the depth of the octree, from 1 to 21. The limiting factor being the Morton code, mentioned in Section~\ref{sec:mapping_framework}, which is specified to be 64-bit. \proposedMappingFramework{} can therefore cover at most $2^{21} \times 0.001$ \SI{}{\metre} = \SI{2097.152}{\metre} in each dimension, with \SI{1}{\milli\metre} voxel size. OctoMap's depth is in the current implementation fixed at 16 depth levels. Meaning, OctoMap can cover at most $2^{16} \times 0.001$ \SI{}{\metre} = \SI{65.536}{\metre} in each dimension, with the same voxel size.

\subsection{Exploration}
\label{sec:exploration}
In the second test, we compare \proposedMappingFramework{} with OctoMap in a next-best-view exploration scenario. We have chosen this scenario since it incorporates all of the above use cases in a realistic setting. We have chosen to use the receding horizon next-best-view exploration method, proposed in \cite{bircher2016receding}, for this comparison. For each node that is being sampled for the RRT, there is a check if the node is in free space. There is also a check if the path between the newly sampled node and prospecting parent node is clear. Lastly, when selecting where to move next we calculate a score for each node in the RRT. This score depends on the distance and the information gain along the branch to the node.

In \proposedMappingFramework{}, as the exploration proceeds the benefit of the proposed approach increases. As more of the space gets classified as free space, all of the above calculations are accelerated in \proposedMappingFramework{} compared to OctoMap. This means that when the environment is almost fully explored you can discard nodes that give very little new information quickly. This shows that \proposedMappingFramework{} is especially well suited for exploration, compared to OctoMap where the necessary calculations are not affected to the same degree by how much is explored.

\begin{figure}[t]
    \centering
    \begin{tikzpicture}
        \begin{axis}[
            xlabel={Time (\SI{}{\second})},
            ylabel={Coverage (\SI{}{\metre\cubed})},
            xmin=0, xmax=900,
            ymin=0, ymax=24000,
            xtick={0,150,300,450,600,750,900},
            ytick={0,4000,8000,12000,16000,20000,24000},
            legend cell align={left},
            legend pos=south east,
            xmajorgrids=true,
            ymajorgrids=true,
            grid style=solid,
        ]
            \addplot[color=blue,thick]
            table[x=time,y=coverage,col sep=space]{ufomap_exploration.csv};
            \addlegendentry{\proposedMappingFramework{}}
            
            \addplot[dashed,color=red,thick]
            table[x=time,y=coverage,col sep=space]{octomap_exploration.csv};
            \addlegendentry{OctoMap}
        \end{axis}
    \end{tikzpicture}
    \caption{Comparision of exploration progress between \proposedMappingFramework{} and OctoMap in the power plant scenario.}
    \label{fig:exploration}
\end{figure}
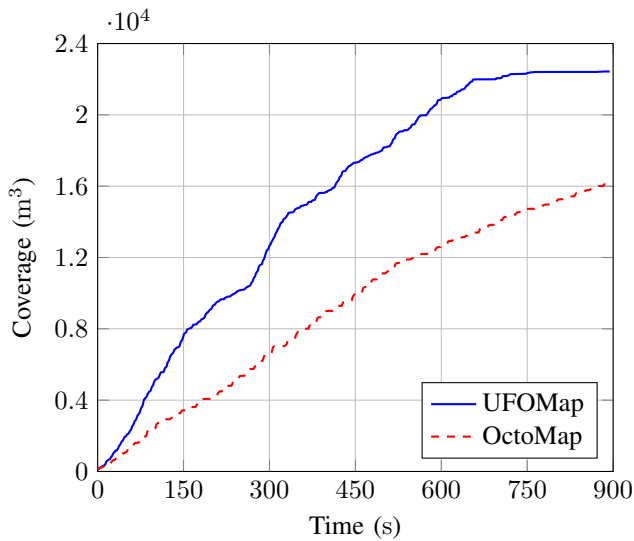

The power plant scenario from the Gazebo model library\footnote{\url{https://bitbucket.org/osrf/gazebo_models/src}} was used for the exploration test, seen in Fig.~\ref{fig:introduction}. \SI{16}{\centi\metre} voxel size was used. As can be seen in Fig.~\ref{fig:exploration},  switching from OctoMap to \proposedMappingFramework{} makes a significant difference to the exploration rate. \proposedMappingFramework{} managed to finish the exploration after around \SI{650}{\second}, while OctoMap had completed 70\% of the exploration after \SI{900}{\second}, which was the maximum allowed time.

We also investigated running the exploration without mutexes, that is, allowing the mapping and the exploration sides simultaneous access to the map. Both OctoMap and \proposedMappingFramework{} crashed. However, \proposedMappingFramework{} did not crash when the indicator $i_{a}$ was used. As described in Section~\ref{sec:multi_threading}, this indicator turns off the automatic pruning, since nodes are not actually removed from the octree in this case.

\section{Conclusion}
\label{sec:conclusion}
We present \proposedMappingFramework{}, an open source framework for 3D mapping. \proposedMappingFramework{} was built on OctoMap, which is one of the basic building blocks in a number of different robotics applications. Just like OctoMap, the underlying data structure is an octree. OctoMap only explicitly models occupied and free space, while \proposedMappingFramework{} explicitly models occupied, free, and unknown space. This representation, together with the fact that every node in the octree stores indicators for what kind of space their children contains, results in a significant performance boost compared to OctoMap for use cases where unknown space is extensively used.

Along with these improvements, we introduce new ways of integrating data into the octree. We show that this leads to significant reductions in the time to insert new measurement data, such as point clouds, into the map.

The \proposedMappingFramework{} mapping framework is freely available at \url{https://github.com/danielduberg/UFOMap} and can be easily integrated with robotics systems. It is written in C++ and can be run as a standalone package or integrated into ROS~\cite{ros}.

\bibliographystyle{IEEEtran}
\bibliography{IEEEabrv,references}

\end{document}